\documentclass[english, letterpaper, 10 pt, conference]{ieeeconf}
\usepackage[T1]{fontenc}
\usepackage[latin9]{inputenc}
\usepackage{amsmath}
\usepackage{cite}
\usepackage[caption=false]{subfig}
\usepackage{dsfont}
\usepackage{bm}
\usepackage{url}

\IEEEoverridecommandlockouts
\makeatletter

\usepackage{tikz}
\usepackage{tikzscale}
\usepackage{amssymb}

\usepackage{enumerate}   
\usepackage{graphicx} 
\usepackage[binary-units=true]{siunitx}
\usepackage{pgfplots}
\pgfplotsset{compat = 1.12}
\usepackage{cleveref}
\usepackage{algorithm}
\usepackage{algorithmicx}
\usepackage{algpseudocode} 
\usepackage{booktabs}
\usepackage{xspace}
\usepackage{balance}
\usetikzlibrary{external}
\usetikzlibrary{matrix}
\usetikzlibrary{pgfplots.groupplots}
\usetikzlibrary{shapes,arrows,positioning,calc}

\newcounter{tagnumb}
\newcommand*{\ev}{{ego}-vehicle\@\xspace}

\newcommand*{\eg}{\textit{e.g.}\@\xspace}

\DeclareMathOperator{\sigm}{sigm}

\crefname{defn}{def.}{defs.}
\Crefname{defn}{Definition}{Definitions}
\crefname{algocf}{alg.}{algs.}
\Crefname{algocf}{Algorithm}{Algorithms}

\newcommand{\includegtikz}[2][]{ 
	\tikzsetnextfilename{#2}
	\includegraphics[#1]{figures/#2} 
}

\makeatother

\usepackage{babel}
\usepackage{diagbox}

\title{\LARGE \bf An LSTM Network for Highway Trajectory Prediction}

\author{Florent Altch\'e$^{2,1}$ and Arnaud de La Fortelle$^{1}$
\thanks{$^{1}$ MINES ParisTech, PSL Research University, Centre for robotics, 60 Bd St Michel 75006 Paris, France {\tt\small [florent.altche, arnaud.de\_la\_fortelle] @mines-paristech.fr}}
\thanks{$^{2}$ \'Ecole des Ponts ParisTech, Cit\'e Descartes, 6-8 Av Blaise Pascal, 77455 Champs-sur-Marne, France}%
}

\begin{document}

\maketitle
\thispagestyle{empty}
\pagestyle{empty}

\begin{abstract}
	In order to drive safely and efficiently on public roads, autonomous vehicles will have to understand the intentions of surrounding vehicles, and adapt their own behavior accordingly. If experienced human drivers are generally good at inferring other vehicles' motion up to a few seconds in the future, most current Advanced Driving Assistance Systems (ADAS) are unable to perform such medium-term forecasts, and are usually limited to high-likelihood situations such as emergency braking. In this article, we present a first step towards consistent trajectory prediction by introducing a long short-term memory (LSTM) neural network, which is capable of accurately predicting future longitudinal and lateral trajectories for vehicles on highway. Unlike previous work focusing on a low number of trajectories collected from a few drivers, our network was trained and validated on the NGSIM US-101 dataset, which contains a total of 800 hours of recorded trajectories in various traffic densities, representing more than 6000 individual drivers.
\end{abstract}

\section{Introduction}
In most situations, experienced human drivers are able to properly infer future behaviors for the surrounding vehicles, which is critical when making tactical driving decisions such as overtaking or crossing an unsignalized intersection. This predictive ability is often lacking in current Advanced Driving Assistance Systems (ADAS) such as Adaptive Cruise Control (ACC), which usually act in a purely reactive fashion and leave tactical decision-making to the driver. On the other hand, fully autonomous vehicles lacking predictive capacities generally have to behave very conservatively in the presence of other traffic participants; as demonstrated by the low-speed collision between a self-driving car and a passenger bus~\cite{Google2016}, reliable motion prediction of surrounding vehicles is a critical feature for safe and efficient autonomous driving.

Many approaches to motion prediction have been proposed in the literature, and a survey can be found in~\cite{Lefevre2014}. As in many machine learning applications, existing techniques can be split between \textit{classification} or \textit{regression} methods. When applied to motion prediction, classification problems consist in determining a high-level behavior (or \textit{intention}), for instance \textit{lane change left}, \textit{lane change right} or \textit{lane keeping} for highway driving or \textit{turn left}, \textit{turn right} or \textit{go straight} in an intersection. Many techniques have already been explored for behavior prediction, such as hidden Markov models~\cite{Tay2012,Streubel2014}, Kalman filtering~\cite{carvalho2014stochastic}, Support Vector Machines~\cite{Mandalia2005,Kumar2013} or directly using a vehicle model~\cite{Houenou2013}; more recently, artificial neural network approaches have also been proposed~\cite{Yoon2016,Khosroshahi2016,Phillips2017}. 

\begin{figure}
	\subfloat[Lateral position]{\includegtikz[width=0.9\columnwidth,height=4.5cm]{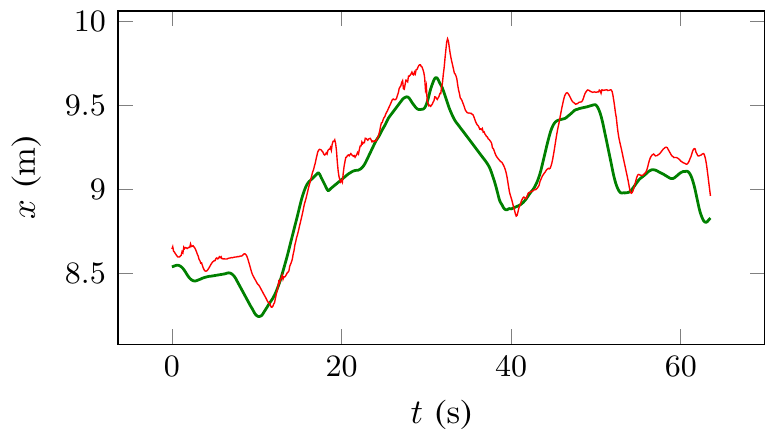}} 
	
	\subfloat[Longitudinal velocity]{\includegtikz[width=0.9\columnwidth,height=4.5cm]{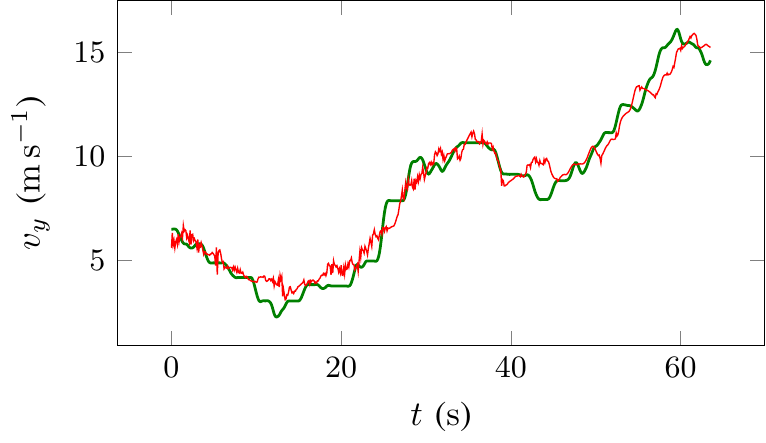}}
	\caption{Example of a predicted trajectory (\SI{2}{\second} forecast) for a vehicle in the test set (in red); the thick green line corresponds to the reference. \label{fig:prediction}}
\end{figure}

The main advantage of predicting behaviors is that the discrete outputs make it easier to train models and evaluate their performance. However, they only provide rough information on future vehicle states, which is not easy to use when planning a trajectory for the self-driving \ev. Some authors have proposed using a generative model, for instance Gaussian Processes~\cite{Tay2012} or neural networks~\cite{Yoon2016} based upon the output of the behavior prediction, but this approach requires multiple trainings and is only as robust as the classifier accuracy. Regression problems, on the other hand, aim at directly obtaining a prediction for future positions of the considered vehicle, which can then be used for motion planning. Many regression algorithms could be used for this problem, for instance regression forests~\cite{Volz2016}; more recently, artificial neural networks have attracted the most attention in the field of trajectory prediction for cars~\cite{Tomar2012,QiangLiu2014}, cyclists~\cite{Zernetsch2016} or pedestrians~\cite{YanjieDuan2016,Alahi2016}. A potential downside of such approaches is that the output of many regression algorithms is a single ``point'' (\eg, a single predicted trajectory) without providing a measure of confidence. To counter this issue, well-established approaches such as Monte Carlo sampling or k-fold validation~\cite{Fushiki2011} can be used to provide error estimates; more recently, dropout estimation techniques have also been proposed for applications using neural networks~\cite{pmlr-v48-gal16}.

In this article, we focus on trajectory prediction using long short-term memory (LSTM) neural networks~\cite{hochreiter1997long}, which are a particular implementation of recurrent neural networks. Because they are able to keep a memory of previous inputs, LSTMs are considered particularly efficient for time series prediction~\cite{Gers2000} and have been widely used in the past few years for pedestrian trajectory prediction~\cite{YanjieDuan2016,Alahi2016} or to predict vehicle destinations at an intersection~\cite{Phillips2017,Zyner2017}. Our main contribution is the design of an LSTM network to predict car trajectories on highways, which is notably critical for safe autonomous overtaking or lane changes, and for which very little literature exists. 

A particular challenge for this problem is that highway driving usually comprises a lot of constant velocity phases with rare punctual events such as lane changes, which are therefore hard to learn correctly. For this reason, many authors rely on purposely recorded~\cite{Kumar2013} or handpicked~\cite{Ding2013,Zheng2014} trajectory sets which are not representative of actual, average driving. Therefore, the real-world performance of trained models can be significantly different. A second contribution of this article is that we train and validate our model using the entire NGSIM US101 dataset~\cite{us101} without \textit{a-priori} selection, and show that we can predict future trajectories with a satisfying average RMS error below \SI{0.7}{\meter} (laterally) and \SI{2.5}{\meter\per\second} (longitudinally) when predicting \SI{10}{\second} ahead. To the best of our knowledge, no other published learning technique was demonstrated with similar results using a dataset representative of real-world conditions.

The rest of this article is structured as follows: in \Cref{sec:problem}, we define the trajectory prediction problem that we are aiming to solve. In \Cref{sec:dataset}, we detail the preprocessing of the US101 dataset to extract the input features of the model, which is presented in \Cref{sec:model}. In \Cref{sec:results}, we present the training procedure and outputs of the trained model. Finally, \Cref{sec:conclusion} concludes the study.

\section{Problem statement\label{sec:problem}}
We consider the problem of predicting future trajectories of vehicles driving on a highway, using previously observed data; these predictions can then be used to plan the motion of an autonomous vehicle.

Formally, we consider a set of observable features $\mathcal I$ and a set of target outputs $\mathcal O$ to be predicted. We assume that the features can all be acquired simultaneously at regular intervals, and that $K_{prev}+1$ successive measurements are always available; we let $T_{prev} = \{-K_{prev}, \dots, 0\}$ and, for $x \in \mathcal I$ and $k \in T_{prev}$, we denote by $x_k$ the value of feature $x$ observed $|k|$ time steps earlier. Similarly, we define $T_{post} = \{0, \dots, K_{post}\}$ and we denote by $y_k$ the value of output $y \in \mathcal O$, $k \in T_{post}$ time steps in the future. We use uppercase $X = (x_k)_{x \in \mathcal I, k \in T_{prev}}$ and $Y = (y_k)_{y \in \mathcal O, k \in T_{post}}$ to respectively denote the tensors of the previously observed features and corresponding predicted outputs. We propose to use a machine learning approach, in which we train a regression function $f$ such that the predicted outputs $\hat Y = f(X)$ match the actual values as closely as possible.

In this article, our approach is to train a predictor for the future trajectory of a single ``target'' vehicle; in order to only use data which can realistically be gathered, we limit the amount of available information to the vehicles immediately around the target vehicle, as described in~\Cref{sec:dataset}. As with many learning approaches, one difficulty is to design models that are able to generalize well from the training data~\cite{Phillips2017}. A second difficulty, more specific to the problem of highway trajectory prediction, is the imbalance between constant velocity driving phases, which are much more frequent than events such as lane changes. 

\section{Data and features\label{sec:dataset}}
\subsection{Dataset}
In this article, we use the Next Generation Simulation (NGSIM) dataset~\cite{us101}, collected in 2005 by the United States Federal Highway Administration, which is one of the largest publicly available source of naturalistic driving data and, as such, has been widely studied in the literature (see, \eg,~\cite{Tomar2012,Yoon2016,Morton2016,Phillips2017}). More specifically, we consider the US101 dataset which contains 45 minutes of trajectories for vehicles on the US101 highway, between 7:50am and 8:35am during the transition from fluid traffic to saturation at rush hour. In total, the dataset contains trajectories for more than 6000 individual vehicles, recorded at \SI{10}{\hertz}.


The NGSIM dataset provides vehicle trajectories in the form of $(X,Y)$ coordinates of the front center of the vehicle in a global frame, and of local $(x,y)$ coordinates of the same point on a road-aligned frame. In this article, we use the local coordinates (dataset columns 5 and 6), where $x$ is the lateral position of the vehicle relative to the leftmost edge of the road, and $y$ its longitudinal position. Moreover, the dataset contains each vehicle's lane identifier at every time step, as well as information on vehicle dimensions and type (motorcycle, car or truck). Finally, the data also contains the identifier of the preceding vehicle for every element in the set (when applicable).

\begin{figure}
	\subfloat[Lateral position]{\includegtikz[width=0.8\columnwidth,height=4cm]{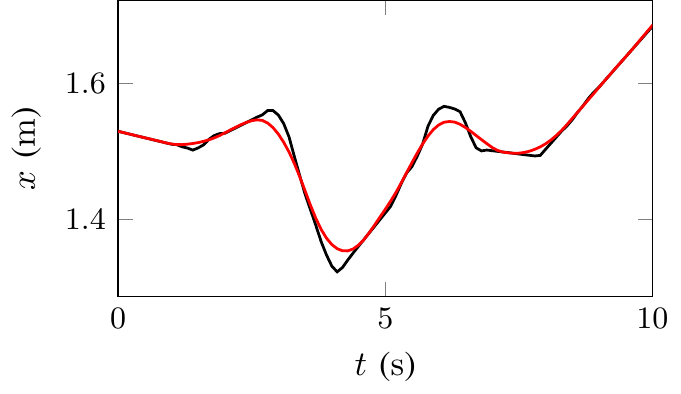}}
	
	\subfloat[Lateral speed]{\includegtikz[width=0.8\columnwidth,height=4cm]{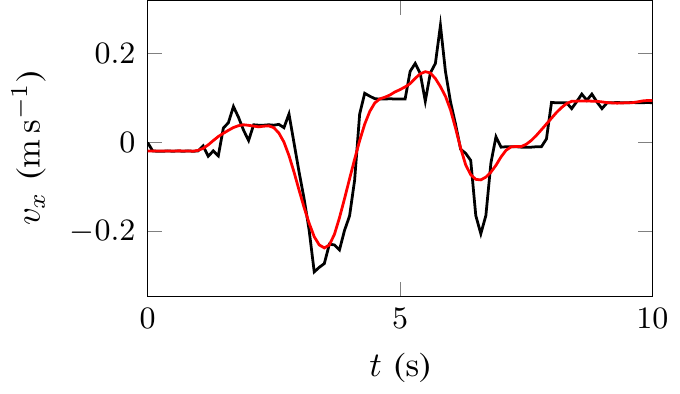}}
	\caption{Smoothing of the lateral position and speed. \label{fig:filtering}}
\end{figure}

\subsection{Data preparation}
One known limitation of the NGSIM set is that vehicle positioning data was obtained from video analysis, and the recorded trajectories contain a significant amount of noise~\cite{montanino2013making}. Velocities, which are obtained from numerical differentiation, suffer even more from this noise. For this reason, we used a first order Savitzky-Golay filter~\cite{savitzky1964smoothing} -- which performs well for signal differentiation -- with window length 11 (corresponding to a time window of \SI{1}{\second}) to smooth the longitudinal and lateral positions and compute the corresponding velocities, as illustrated in \Cref{fig:filtering}.

\begin{figure}
	\centering \includegtikz[width=0.8\columnwidth]{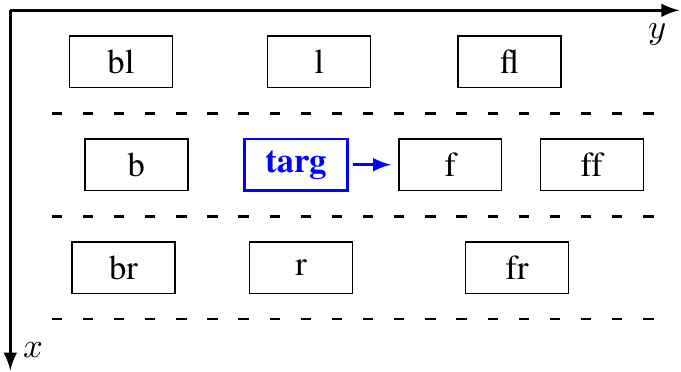}
	\caption{Vehicles of interest around the target vehicle and local axis system. The blue arrow represents traffic direction. \label{fig:vehicles}}
\end{figure}

In this article, we hypothesize that the future behavior of a target vehicle can be reliably predicted by using local information on the vehicles immediately around it; a similar hypothesis was successfully tested in~\cite{Schlechtriemen2014} to detect lane-change intent. For a target vehicle, we consider 9 vehicles of interest, that we label according to their relative position with respect to the target vehicle $targ$, as shown in \Cref{fig:vehicles}. By convention, we let $r$ (respectively $l$) be the vehicle which is closest to the target vehicle in a different lane with $x > x_{targ}$ (respectively $x < x_{targ}$). We respectively denote by $fl$, $f$, $fr$ and $ff$ the vehicle preceding $l$, $targ$, $r$ and $f$; similarly, vehicles $bl$, $b$ and $br$ are chosen so that their leader is respectively $l$, $targ$ and $r$. During the data preprocessing phase, we compute the identifier of each vehicle of interest and perform join requests to append their information to the dataset. When such a vehicle does not exist, the corresponding data columns are set to zero. 

Note that the rationale behind the inclusion of information on $ff$ is that only observing the state of the vehicle directly in front is not always sufficient to correctly determine future traffic evolution. For instance, in a jam, knowing that vehicle $ff$ is accelerating can help infer that $f$, although currently stopped, will likely accelerate in the future instead of remaining stopped. The obvious limit to increasing the number of considered vehicles is the ability to realistically gather sufficient data using on-board sensors; for this reason, we restrict the available information to these 9 vehicles..

\subsection{Features}
In this article, we aim at only using features which can be reasonably easily measured using on-board sensors such as GNSS and LiDAR, barring range or occlusion issues. For this reason, we consider a different set of features for the target vehicle (for which we want to compute the future trajectory) and for its surrounding vehicles as described above.

For the target vehicle, we define the following features:
\begin{itemize}
	\item local lateral position $x_{targ}$, to account for different behaviors depending on the driving lane,
	\item local longitudinal position $y_{targ}$, to account for different behaviors when approaching the merging lane,
	\item lateral and longitudinal velocities ${v_x}_{targ}$ and ${v_y}_{targ}$,
	\item type (motorcycle, car or truck), encoded respectively as $-1$, $0$ or $+1$.
\end{itemize}

For each vehicle $p \in \{bl, b, br, l, f, r, fl, f, fr, ff\}$, we define the following features:
\begin{itemize}
	\item lateral velocity ${v_x}_p$,
	\item longitudinal velocity relative to $targ$: $\Delta {v_y}_p = {v_y}_{targ} - {v_y}_p$,
	\item lateral distance from $targ$: $\Delta x_p = x_p - x_{targ}$,
	\item longitudinal distance from $targ$: $\Delta y_p = y_p - y_{targ}$,
	\item signed time-to-collision with $targ$: $TTC_p = \frac{\Delta y_p}{\Delta {v_y}_p}$,
	\item type (motorcycle, car or truck), encoded respectively as $-1$, $0$ or $+1$.
\end{itemize}
These features are scaled to remain in an acceptable range with respect to the activation functions; in this article, we simply divide longitudinal and lateral distances (expressed in SI units), as well as longitudinal velocities by $10$, which results in values generally contained within $[-2,2]$. Note that in the case of missing data (\eg, when the left vehicle does not exist), the corresponding values of $\Delta$ can become higher (in absolute value).

This choice of features was made to replicate the information a human driver is likely to base its decisions upon: the features from surrounding vehicles are all relative to the target vehicle, as we expect drivers to usually make decisions based on perceived distances and relative speeds rather than their values in an absolute frame. Features regarding the target vehicle's speed are given in a (road-relative) absolute frame as drivers are generally aware of speedometer information; similarly, we use road-relative positions since the driver is usually able to visually measure lateral distances from the side of the road, and knows its longitudinal position. The choice of explicitly including time-to-collision as a feature comes from the high importance of this metrics in lane-change decisions~\cite{Schlechtriemen2015}; furthermore, neurosciences seem to indicate that animal and human brains heavily rely on time-to-collision estimations to perform motor tasks~\cite{field2005perceiving}. 

\subsection{Outputs}
In this article, our goal is to predict the future trajectory of the target vehicle. Since the region of interest spans roughly \SI{1}{\kilo\meter} longitudinally, the values of the longitudinal position can become quite large; for this reason, we prefer to predict future longitudinal velocities ${{}\hat v_y}_{targ}$ instead. Since the lateral position is bounded, we directly use ${\hat x}_{targ}$ for the output. In order to have different horizons of prediction, we choose a vector of outputs $[{\hat x}_{targ}^k, {{}\hat v_y}_{targ}^k]_{k=1\dots K}$ consisting in values taken $k$ seconds in the future.

\section{Learning model\label{sec:model}}
\begin{figure}
	\centering \includegtikz[width=\columnwidth]{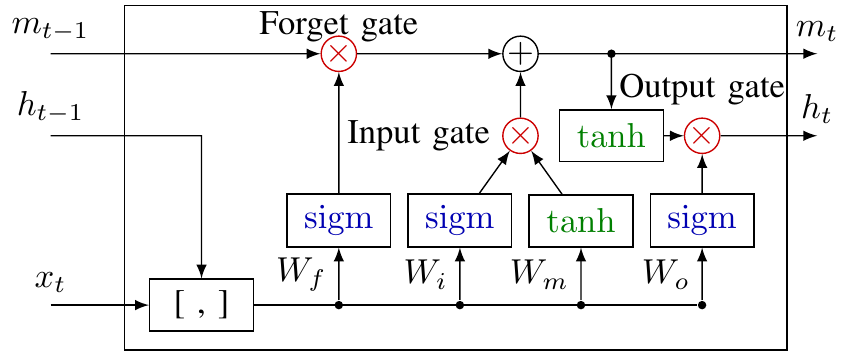}
	\caption{Internal structure of an LSTM cell as used in the Keras framework. $\sigm$ and $\tanh$ respectively denote a sigmoid and hyperbolic tangent activation functions; the $[\ ,\ ]$ node on the lower right operates a concatenation of the new input $x_t$ and the previous output $h_{t-1}$. \label{fig:lstm}}
\end{figure}

Contrary to many existing frameworks for intent or behavior prediction, which can be modeled as \textit{classification} problems, our aim is to predict future $(x,y)$ positions for the target vehicle, which intrinsically is a \textit{regression} problem. Due to their success in many applications, we choose to use an artificial neural network for our learning architecture, in the form of a Long Short-Term Memory (LSTM) network~\cite{hochreiter1997long}. LSTMs are a particular implementation of recurrent neural networks (RNN), which are particularly well suited for time series; in this article, we used the Keras framework~\cite{chollet2015keras}, which implements the extended LSTM described in~\cite{Gers2000}, presented in \Cref{fig:lstm}. Compared to simpler \textit{vanilla} RNN implementations, LSTMs are generally considered more robust for long time series~\cite{hochreiter1997long}; future work will focus on comparing the performance of different RNN approaches on our particular dataset.

\let\OrgPgfTransformScale\pgftransformscale
\renewcommand*{\pgftransformscale}[1]{%
	\gdef\ScaleFactor{#1}%
	\OrgPgfTransformScale{#1}%
}
\def\ScaleFactor{1}

An interesting feature of LSTM cells is the presence of an internal state which serves as the cell's memory, denoted by $m_t$ in \Cref{fig:lstm}. Based on a new input $x_t$, its previous state $m_{t-1}$ and previous output $h_{t-1}$, the cell performs different operations using so-called ``gates'':
\begin{itemize}
	\item forget: uses the inputs to decide how much to ``forget'' from the cell's previous internal state $m_{t-1}$;
	\item input: decides the amount of new information to be stored in memory based on $x_t$ and $h_{t-1}$;
	\item output: computes the new cell output from a mix of the previous states and output of the input gate.
\end{itemize}
This particular feature of LSTMs allows a network to learn long-term relations between features, which makes them very powerful for time series prediction. 

\begin{figure}
	\centering \includegtikz{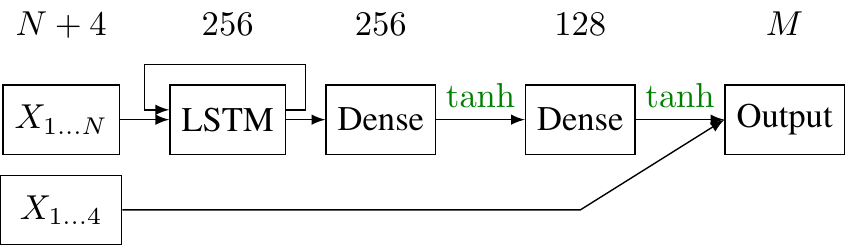}
	\caption{Network architecture used as reference design. The four repeated inputs $X_{1\dots 4}$  correspond to the current target vehicle states (positions and speeds), and are directly fed to the (dense) output layer. \label{fig:network}}
\end{figure}

Due to their recurrent nature, even a single layer of LSTM nodes can be considered as a ``deep'' neural network. Although such layers may theoretically be stacked in a fashion similar to convolutional neural networks to learn higher-level features, previous studies~\cite{Phillips2017} and our own experiments (see \Cref{sec:results}) seem to indicate that stacked layers of LSTM do not provide improvements over a single layer in our application. In this article, we use the network presented in \Cref{fig:network} as our reference architecture, and we compare a few variations on this design in \Cref{sec:results}. The reference architecture uses a first layer of 256 LSTM cells, followed by two dense (fully connected) and time-distributed layers of 256 and 128 neurons and a final dense output layer containing as many cells as the number of outputs. In this simple architecture, the role of the LSTM layer is to abstract a meaningful representation of the input time series; these higher-level ``features'' are then combined by the two dense layers in order to produce the output, in this case the predicted future states.

Additionally, the first four input features of the network -- corresponding to the absolute state of the target vehicle -- are repeated and directly fed to the (dense) output layer, thus bypassing the LSTMs. The motivation behind this bypass is to allow the recurrent layer to focus on variations from the current states, rather than modeling the steady state of driving at constant speed on a given lane. In practice (see \Cref{sec:results}), the use of this bypass seems to slightly improve prediction quality.

\section{Results\label{sec:results}}
In this section, we use the previously described deep neural network to predict future trajectories sampled from the US101 dataset. To assess the learning performance of the model and its ability to generalize over different drivers, we first randomly select 80\% of vehicles (4892 trajectories) for training, and withhold the remaining 20\% of vehicles (1209 trajectories) for testing; these later 20\% are not used during the training phase.

In this article, we aim at designing a network which is capable of understanding medium-term (up to \SI{10}{\second}) relations for prediction. To avoid backpropagation-related issues that can arise with long time series, we trained the network using windows of 100 inputs, representing a total of \SI{10}{\second} past observations. One such window is taken every 10 data points; therefore, two consecutive windows have \SI{9}{\second} of overlap. Additionally, vehicles are grouped by batches of 500 (except for the final batch), and data is shuffled within batches. As a result, the data actually fed to the network for a batch of vehicles is a tridimensional tensor of shape $B \times 100 \times N$ where $B \approx 20000$ and $N \approx 50$ are respectively the total number of time windows in the batch, and the number of features. The training is performed on GPU using the TensorFlow backend with a batch size of 32; the model is trained for 5 epochs on each set of 500 vehicles and the whole dataset is processed 20 times, resulting in 100 effective epochs.

For the test set, we directly feed the input features for the whole trajectory, without processing the data by time windows. For each vehicle, we then compute the Root Mean Squared error (RMSE) between the network prediction and the actual expected value. In \Cref{fig:prediction}, we present the prediction outputs of the network of \Cref{fig:network} for one of the vehicles in the test set. For comparison purposes, we tested the following variations of the reference design:
\begin{itemize}
	\item Reference design of \Cref{fig:network},
	\item Using vehicle type information,
	\item Without using information on vehicle $ff$,
	\item Without using a bypass,
	\item Using bypass before the first dense layer (only bypass the LSTMs),
	\item Using a linear activation for the $128$ dense layer,
	\item Adding another LSTM layer after the first,
	\item Adding a third dense layer of $64$ nodes;
\end{itemize}
\Cref{tab:rmse-allnets} presents the average RMS error across all networks for various prediction horizons. In an effort to further improve accuracy, we used a light bagging technique consisting in using the average of the outputs from the four best models (denoted by a * in \Cref{tab:rmse-allnets}); this bagged predictor almost always perform best over the testing data. For comparison purposes, we also report results from~\cite{QiangLiu2014} which chose a related approach using a multi-layer perceptron (which does not have a recurrent layer). The higher prediction errors for longer horizons seem to show that the use of LSTMs provides better results for longer prediction horizons.

\begin{table}
	\centering \caption{RMS error for the tested models}\label{tab:rmse-allnets}
	\subfloat[Lateral position (errors are in \SI{}{\meter})]{
		\begin{tabular}{l c c c c c c c}
		& \multicolumn{7}{c}{Prediction horizon} \\
		\cmidrule{2-8}
		Model & \SI{1}{\second} & \SI{2}{\second} & \SI{3}{\second} & \SI{4}{\second} &  \SI{6}{\second} & \SI{8}{\second} &  \SI{10}{\second} \\
		\midrule
		Reference* & \textbf{0.11} & \textbf{0.25} & \textbf{0.33} & \textbf{0.40} & {0.53} & {0.60} & 0.73 \\
		Type* & 0.39 & 0.39 & 0.44 & 0.48 & {0.53} & 0.63 & {0.69} \\
		No $ff$* & 0.14 & 0.24 & 0.33 & 0.41 & 0.54 & 0.65 & 0.76 \\
		No bypass & 0.80 & 0.82 & 0.85 & 0.88 & 0.93 & 0.97 & 1.03 \\
		Bypass before & 0.33 & 0.38 & 0.43 & 0.46 & 0.52 & 0.61 & 0.68 \\
		Lin. activ. & 1.38 & 1.39 & 1.40 & 1.42 & 1.46 & 1.51 & 1.56 \\
		2 LSTMs & 1.25 & 1.26 & 1.28 & 1.29 & 1.33 & 1.37 & 1.41 \\
		3 dense* & 0.34 & 0.38 & 0.44 & 0.50 & 0.59 & 0.70 & 0.72 \\
		\cite{QiangLiu2014} & \textbf{0.11} & 0.32 & 0.71 & \multicolumn{4}{c}{not available} \\
		\midrule
		Bagged & 0.17 & \textbf{0.25} & \textbf{0.33} & \textbf{0.40} & \textbf{0.46} & \textbf{0.57} & \textbf{0.65} \\
		\bottomrule
	\end{tabular}
	}

\subfloat[Longitudinal speed (errors are in \SI{}{\meter\per\second})]{
	\begin{tabular}{l c c c c c c c}
		& \multicolumn{7}{c}{Prediction horizon} \\
		\cmidrule{2-8}
		Model & \SI{1}{\second} & \SI{2}{\second} & \SI{3}{\second} & \SI{4}{\second} &  \SI{6}{\second} & \SI{8}{\second} &  \SI{10}{\second} \\
		\midrule
	Reference* & 0.71 & 0.99 & 1.25 & 1.49 & 2.10 & 2.60 & 2.96 \\
	Type* & {0.65} & 0.88 & 1.05 & \textbf{1.25} & \textbf{1.75} & \textbf{2.28} & \textbf{2.74} \\
	No $ff$* & 0.67 & 0.91 & 1.16 & 1.44 & 1.98 & 2.43 & 2.84 \\
	No bypass &1.50 & 1.50 & 1.55 & 1.66 & 2.05 & 2.50 & 2.89 \\
	Bypass before & 0.78 & 0.90 & 1.06 & 1.26 & 1.76 & 2.30 & 2.78 \\
	Lin. activ. & 0.77 & 1.10 & 1.34 & 1.56 & 2.08 & 2.58 & 2.94 \\
	2 LSTMs & 0.76 & 1.14 & 1.42 & 1.71 & 2.22 & 2.72 & 3.17 \\
	3 dense* & 0.73 & \textbf{0.87} & \textbf{1.04} & \textbf{1.25} & 1.76 & 2.30 & 2.77 \\
	\midrule
	Bagged & \textbf{0.64} & \textbf{0.81} & \textbf{0.98} & \textbf{1.18} & \textbf{1.63} & \textbf{2.08} & \textbf{2.48} \\
	\bottomrule
\end{tabular}
}
	
\end{table}

\begin{figure}
	\centering \subfloat[Lateral position]{	\includegtikz[width=\columnwidth,height=5cm]{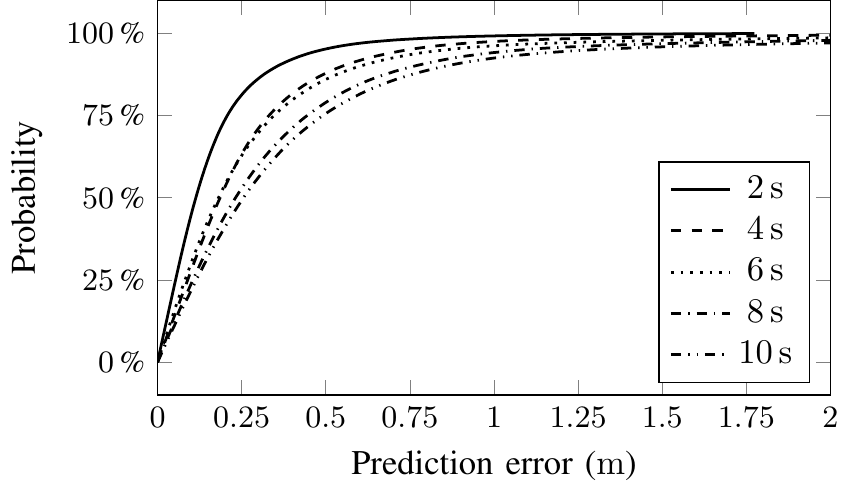}}
	
	\subfloat[Longitudinal velocity]{		\includegtikz[width=\columnwidth,height=5cm]{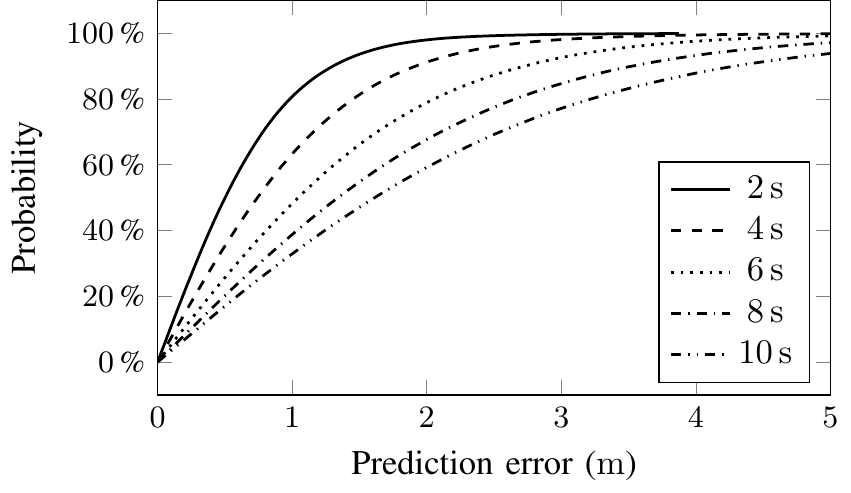}}
	\caption{Distribution of error on the test set for the bagged predictor. \label{fig:distribution-n1}}
\end{figure}

As can be seen in \Cref{tab:rmse-allnets}, the architecture of \Cref{fig:network} provides the best overall results for lateral position prediction, but is less precise for velocity prediction. Interestingly, providing vehicle type information does not improve predictions of lateral movement but allows more precise forecasting of longitudinal speed, probably due to the difference in acceleration capacities. In what follows, we focus on this reference design to provide more insight on error characterization. \Cref{fig:distribution-n1} presents the distribution of prediction error over the test set for the bagged predictor.

\begin{figure}
	\centering \subfloat[Lateral position]{\includegtikz[width=0.9\columnwidth,height=4cm]{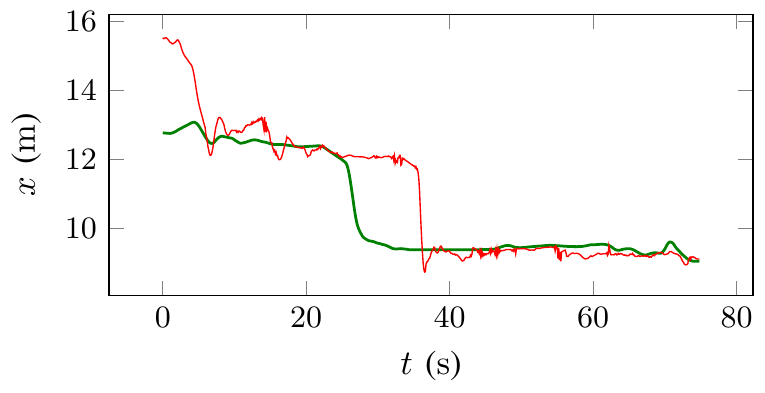}}
	
	\centering \subfloat[Longitudinal speed]{\includegtikz[width=0.9\columnwidth,height=4cm]{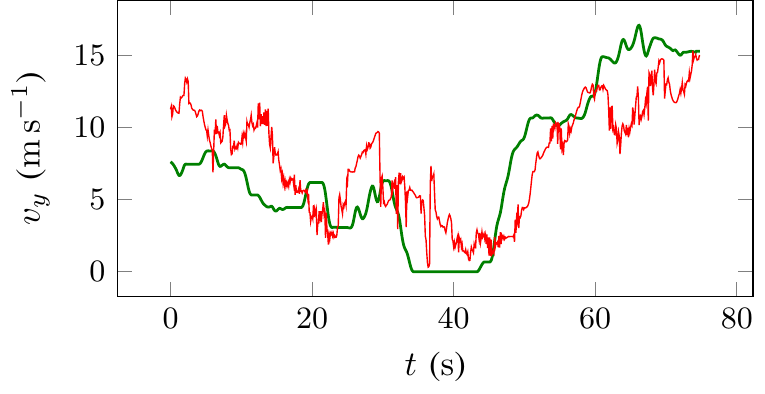}}
	\caption{Delay between prediction (in red) and reference (in green, thick line) for a prediction horizon of \SI{10}{\second}. Note that although a large delay is observed on lateral position prediction, it is much smaller for longitudinal speed. \label{fig:prediction-delay}}
\end{figure}

Note that the above results mostly use the RMSE and error distributions to evaluate the quality of prediction. However, such aggregated metrics may not be the best suited for this particular application, notably due to the over-representation of straight driving at constant speed, which highly outnumber discrete events such as lane changes or sudden acceleration. An illustration of this limitation is that we sometimes observe that the prediction reacts with a delay, such as shown in \Cref{fig:prediction-delay}; this effect mostly happens for longer prediction horizons, and is not properly accounted for using RMSE. In the worst cases (such as depicted in \Cref{fig:prediction-delay}), this delay can reach up to \SI{8}{\second} or \SI{9}{\second} for a prediction horizon of \SI{10}{\second}, thus demonstrating that the model is sometimes unable to interpret observed behaviors.

Experimentally, separately training each network output seems to yield better results, at the cost of an increased overall training time; training one model per vehicle type, or using wider networks could also be possible ways of improvement, as well as using different time windows durations for training. Besides providing improvement to the model, future work will focus on designing better suited metrics related to correct detection of meaningful traffic information, for instance lane changes, overtaking events or re-acceleration and braking during stop-start driving, which could help further improve predictions. Moreover, a more careful analysis of cases showing large deviations should be performed to compare model predictions with human-made estimations.

\section{Conclusion\label{sec:conclusion}}
In this article, we proposed a neural network architecture based on LSTMs to predict future vehicle trajectories on highway using naturalistic driving data from the widely studied NGSIM dataset. This network was shown to achieve better prediction accuracy than the previous state-of-the-art, with an average RMS error of roughly \SI{70}{\centi\meter} for the lateral position \SI{10}{\second} in the future, and lower than \SI{3}{\meter\per\second} for the longitudinal velocity with the same horizon. Contrary to many previous studies which used handpicked trajectories for training and testing, thus adding a selection bias, our results were obtained using the whole US101 dataset, which should make the model more apt to deal with real-world scenarios.

Although this work is highly preliminary and some limitations -- notably the observed delayed response -- should be addressed, we believe that these results constitute a promising basis to compute probable trajectories for surrounding vehicles. The use of the actual predictions alongside with precise statistics on error distribution could, in turn, be used to significantly improve current motion planning algorithms. Provided the discussed limitations can be overcome, our results open many perspectives for future research, first by studying their generalizability to other highways, then other driving scenarios such as in intersections and roundabouts. To this end, a proper study of selected features should be performed, both to determine a good inputs scaling technique and to study which features are the most relevant. Moreover, more in-depth investigation of error distribution using application-specific metrics should be performed to properly validate the proposed models. Finally, our current approach does not consider stochasticity or confidence levels, although this information is essential to correctly use the predictions; future work will investigate possible techniques to output probability distributions instead of single values.

\bibliographystyle{IEEEtranurl}
\balance
\bibliography{trajprediction}

\begin{thebibliography}{10}
\providecommand{\url}[1]{#1}
\csname url@rmstyle\endcsname
\providecommand{\newblock}{\relax}
\providecommand{\bibinfo}[2]{#2}
\providecommand\BIBentrySTDinterwordspacing{\spaceskip=0pt\relax}
\providecommand\BIBentryALTinterwordstretchfactor{4}
\providecommand\BIBentryALTinterwordspacing{\spaceskip=\fontdimen2\font plus
\BIBentryALTinterwordstretchfactor\fontdimen3\font minus
  \fontdimen4\font\relax}
\providecommand\BIBforeignlanguage[2]{{%
\expandafter\ifx\csname l@#1\endcsname\relax
\typeout{** WARNING: IEEEtran.bst: No hyphenation pattern has been}%
\typeout{** loaded for the language `#1'. Using the pattern for}%
\typeout{** the default language instead.}%
\else
\language=\csname l@#1\endcsname
\fi
#2}}

\bibitem{Google2016}
Google, ``Google self-driving car project monthly report,''
  https://www.google.com/selfdrivingcar/files/reports/report-0216.pdf, Tech.
  Rep., Feb. 2016.

\bibitem{Lefevre2014}
\BIBentryALTinterwordspacing
S.~Lef{\`{e}}vre, D.~Vasquez, and C.~Laugier, ``{A survey on motion prediction
  and risk assessment for intelligent vehicles},'' \emph{ROBOMECH Journal},
  vol.~1, no.~1, p.~1, dec 2014.
\BIBentrySTDinterwordspacing

\bibitem{Tay2012}
\BIBentryALTinterwordspacing
C.~Tay, K.~Mekhnacha, and C.~Laugier, ``\BIBforeignlanguage{en}{{Probabilistic
  Vehicle Motion Modeling and Risk Estimation}},'' in
  \emph{\BIBforeignlanguage{en}{Handbook of Intelligent Vehicles}}.\hskip 1em
  plus 0.5em minus 0.4em\relax Springer London, 2012, pp. 1479--1516.
\BIBentrySTDinterwordspacing

\bibitem{Streubel2014}
T.~Streubel and K.~H. Hoffmann, ``{Prediction of driver intended path at
  intersections},'' \emph{IEEE Intelligent Vehicles Symposium, Proceedings},
  pp. 134--139, 2014.

\bibitem{carvalho2014stochastic}
A.~Carvalho, Y.~Gao, S.~Lefevre, and F.~Borrelli, ``Stochastic predictive
  control of autonomous vehicles in uncertain environments,'' in \emph{12th
  International Symposium on Advanced Vehicle Control}, 2014.

\bibitem{Mandalia2005}
\BIBentryALTinterwordspacing
H.~M. Mandalia and M.~D.~D. Salvucci, ``{Using Support Vector Machines for
  Lane-Change Detection},'' \emph{Proceedings of the Human Factors and
  Ergonomics Society Annual Meeting}, vol.~49, no.~22, pp. 1965--1969, sep
  2005.
\BIBentrySTDinterwordspacing

\bibitem{Kumar2013}
\BIBentryALTinterwordspacing
P.~Kumar, M.~Perrollaz, S.~Lefevre, and C.~Laugier, ``{Learning-based approach
  for online lane change intention prediction},'' in \emph{2013 IEEE
  Intelligent Vehicles Symposium (IV)}.\hskip 1em plus 0.5em minus 0.4em\relax
  IEEE, jun 2013, pp. 797--802.
\BIBentrySTDinterwordspacing

\bibitem{Houenou2013}
\BIBentryALTinterwordspacing
A.~Houenou, P.~Bonnifait, V.~Cherfaoui, and {Wen Yao}, ``{Vehicle trajectory
  prediction based on motion model and maneuver recognition},'' in \emph{2013
  IEEE/RSJ International Conference on Intelligent Robots and Systems}.\hskip
  1em plus 0.5em minus 0.4em\relax IEEE, nov 2013, pp. 4363--4369.
\BIBentrySTDinterwordspacing

\bibitem{Yoon2016}
\BIBentryALTinterwordspacing
S.~Yoon and D.~Kum, ``{The multilayer perceptron approach to lateral motion
  prediction of surrounding vehicles for autonomous vehicles},'' in \emph{2016
  IEEE Intelligent Vehicles Symposium (IV)}, vol. 2016-August.\hskip 1em plus
  0.5em minus 0.4em\relax IEEE, jun 2016, pp. 1307--1312.
\BIBentrySTDinterwordspacing

\bibitem{Khosroshahi2016}
\BIBentryALTinterwordspacing
A.~Khosroshahi, E.~Ohn-Bar, and M.~M. Trivedi, ``{Surround vehicles trajectory
  analysis with recurrent neural networks},'' in \emph{2016 IEEE 19th
  International Conference on Intelligent Transportation Systems (ITSC)}.\hskip
  1em plus 0.5em minus 0.4em\relax IEEE, nov 2016, pp. 2267--2272.
\BIBentrySTDinterwordspacing

\bibitem{Phillips2017}
D.~J. Phillips, T.~A. Wheeler, and M.~J. Kochenderfer, ``{Generalizable
  Intention Prediction of Human Drivers at Intersections},'' \emph{2017 IEEE
  Intelligent Vehicles Symposium (IV)}, pp. 1665--1670, 2017.

\bibitem{Volz2016}
\BIBentryALTinterwordspacing
B.~Volz, H.~Mielenz, R.~Siegwart, and J.~Nieto, ``{Predicting pedestrian
  crossing using Quantile Regression forests},'' in \emph{2016 IEEE Intelligent
  Vehicles Symposium}.\hskip 1em plus 0.5em minus 0.4em\relax IEEE, jun 2016,
  pp. 426--432.
\BIBentrySTDinterwordspacing

\bibitem{Tomar2012}
\BIBentryALTinterwordspacing
R.~S. Tomar and S.~Verma, ``{Safety of Lane Change Maneuver Through A Priori
  Prediction of Trajectory Using Neural Networks},'' \emph{Network Protocols
  and Algorithms}, vol.~4, no.~1, pp. 4--21, 2012.
\BIBentrySTDinterwordspacing

\bibitem{QiangLiu2014}
\BIBentryALTinterwordspacing
{Qiang Liu}, B.~Lathrop, and V.~Butakov, ``{Vehicle lateral position
  prediction: A small step towards a comprehensive risk assessment system},''
  in \emph{17th International IEEE Conference on Intelligent Transportation
  Systems (ITSC)}.\hskip 1em plus 0.5em minus 0.4em\relax IEEE, oct 2014, pp.
  667--672.
\BIBentrySTDinterwordspacing

\bibitem{Zernetsch2016}
\BIBentryALTinterwordspacing
S.~Zernetsch, S.~Kohnen, M.~Goldhammer, K.~Doll, and B.~Sick, ``{Trajectory
  prediction of cyclists using a physical model and an artificial neural
  network},'' in \emph{2016 IEEE Intelligent Vehicles Symposium (IV)}.\hskip
  1em plus 0.5em minus 0.4em\relax IEEE, jun 2016, pp. 833--838.
\BIBentrySTDinterwordspacing

\bibitem{YanjieDuan2016}
\BIBentryALTinterwordspacing
{Yanjie Duan}, {Yisheng Lv}, and {Fei-Yue Wang}, ``{Travel time prediction with
  LSTM neural network},'' in \emph{2016 IEEE 19th International Conference on
  Intelligent Transportation Systems (ITSC)}.\hskip 1em plus 0.5em minus
  0.4em\relax IEEE, nov 2016, pp. 1053--1058.
\BIBentrySTDinterwordspacing

\bibitem{Alahi2016}
\BIBentryALTinterwordspacing
A.~Alahi, K.~Goel, V.~Ramanathan, A.~Robicquet, L.~Fei-Fei, and S.~Savarese,
  ``{Social LSTM: Human Trajectory Prediction in Crowded Spaces},'' in
  \emph{2016 IEEE Conference on Computer Vision and Pattern Recognition
  (CVPR)}.\hskip 1em plus 0.5em minus 0.4em\relax IEEE, jun 2016, pp. 961--971.
\BIBentrySTDinterwordspacing

\bibitem{Fushiki2011}
\BIBentryALTinterwordspacing
T.~Fushiki, ``Estimation of prediction error by using k-fold
  cross-validation,'' \emph{Statistics and Computing}, vol.~21, no.~2, pp.
  137--146, 2011.
\BIBentrySTDinterwordspacing

\bibitem{pmlr-v48-gal16}
\BIBentryALTinterwordspacing
Y.~Gal and Z.~Ghahramani, ``Dropout as a bayesian approximation: Representing
  model uncertainty in deep learning,'' in \emph{Proceedings of The 33rd
  International Conference on Machine Learning}, ser. Proceedings of Machine
  Learning Research, vol.~48.\hskip 1em plus 0.5em minus 0.4em\relax PMLR,
  20--22 Jun 2016, pp. 1050--1059.
\BIBentrySTDinterwordspacing

\bibitem{hochreiter1997long}
S.~Hochreiter and J.~Schmidhuber, ``Long short-term memory,'' \emph{Neural
  computation}, vol.~9, no.~8, pp. 1735--1780, 1997.

\bibitem{Gers2000}
\BIBentryALTinterwordspacing
F.~A. Gers, J.~Schmidhuber, and F.~Cummins, ``{Learning to Forget: Continual
  Prediction with LSTM},'' \emph{Neural Computation}, vol.~12, no.~10, pp.
  2451--2471, oct 2000.
\BIBentrySTDinterwordspacing

\bibitem{Zyner2017}
A.~Zyner, S.~Worrall, J.~Ward, and E.~Nebot, ``{Long Short Term Memory for
  Driver Intent Prediction},'' \emph{2017 IEEE Intelligent Vehicles Symposium
  (IV)}, pp. 1484--1489, 2017.

\bibitem{Ding2013}
\BIBentryALTinterwordspacing
C.~Ding, W.~Wang, X.~Wang, and M.~Baumann, ``{A neural network model for
  driver's lane-changing trajectory prediction in urban traffic flow},''
  \emph{Mathematical Problems in Engineering}, vol. 2013, 2013.
\BIBentrySTDinterwordspacing

\bibitem{Zheng2014}
\BIBentryALTinterwordspacing
J.~Zheng, K.~Suzuki, and M.~Fujita, ``{Predicting driver's lane-changing
  decisions using a neural network model},'' \emph{Simulation Modelling
  Practice and Theory}, vol.~42, pp. 73--83, 2014.
\BIBentrySTDinterwordspacing

\bibitem{us101}
\BIBentryALTinterwordspacing
{U.S. Federal Highway Administration}. (2005) {US} {H}ighway 101 dataset.
\BIBentrySTDinterwordspacing

\bibitem{Morton2016}
J.~Morton and T.~A. Wheeler, ``{Project Report Deep Learning of Spatial and
  Temporal Features for Automotive Prediction},'' pp. 1--9, 2016.

\bibitem{montanino2013making}
M.~Montanino and V.~Punzo, ``Making ngsim data usable for studies on traffic
  flow theory: Multistep method for vehicle trajectory reconstruction,''
  \emph{Transportation Research Record: Journal of the Transportation Research
  Board}, no. 2390, pp. 99--111, 2013.

\bibitem{savitzky1964smoothing}
A.~Savitzky and M.~J. Golay, ``Smoothing and differentiation of data by
  simplified least squares procedures.'' \emph{Analytical chemistry}, vol.~36,
  no.~8, pp. 1627--1639, 1964.

\bibitem{Schlechtriemen2014}
\BIBentryALTinterwordspacing
J.~Schlechtriemen, A.~Wedel, J.~Hillenbrand, G.~Breuel, and K.-d. Kuhnert, ``{A
  lane change detection approach using feature ranking with maximized
  predictive power},'' in \emph{2014 IEEE Intelligent Vehicles Symposium
  Proceedings}.\hskip 1em plus 0.5em minus 0.4em\relax IEEE, jun 2014, pp.
  108--114.
\BIBentrySTDinterwordspacing

\bibitem{Schlechtriemen2015}
J.~Schlechtriemen, F.~Wirthmueller, A.~Wedel, G.~Breuel, and K.~D. Kuhnert,
  ``{When will it change the lane? A probabilistic regression approach for
  rarely occurring events},'' \emph{IEEE Intelligent Vehicles Symposium,
  Proceedings}, vol. 2015-August, pp. 1373--1379, 2015.

\bibitem{field2005perceiving}
D.~T. Field and J.~P. Wann, ``Perceiving time to collision activates the
  sensorimotor cortex,'' \emph{Current Biology}, vol.~15, no.~5, pp. 453--458,
  2005.

\bibitem{chollet2015keras}
F.~Chollet \emph{et~al.}, ``Keras,'' \url{https://github.com/fchollet/keras},
  2015.

\end{thebibliography}

\end{document}